\documentclass[10pt,twocolumn,letterpaper]{article}

\usepackage[pagenumbers]{cvpr} 

\makeatletter
\@namedef{ver@everyshi.sty}{}
\makeatother
\usepackage{tikz}

\usepackage{graphicx}
\usepackage{amsmath}
\usepackage{amssymb}
\usepackage{booktabs}
\usepackage{mmstyle}
\usepackage{multirow}

\usepackage{times}
\usepackage{epsfig}
\usepackage{bm}
\usepackage{color}
\usepackage{colortbl}
\usepackage{pgfplots}
\usepackage[misc]{ifsym}
\usepackage{lipsum}

\newcommand\blfootnote[1]{%
  \begingroup
  \renewcommand\thefootnote{}\footnote{#1}%
  \addtocounter{footnote}{-1}%
  \endgroup
}

\usepackage[pagebackref,breaklinks,colorlinks]{hyperref}

\usepackage[capitalize]{cleveref}
\crefname{section}{Sec.}{Secs.}
\Crefname{section}{Section}{Sections}
\Crefname{table}{Table}{Tables}
\crefname{table}{Tab.}{Tabs.}

\def\cvprPaperID{3930} 
\def\confName{CVPR}
\def\confYear{2022}
\definecolor{mygray}{gray}{0.95}
\definecolor{myred}{rgb}{1.0, 0.0, 0.0}

\newcommand{\tablestyle}[2]{\setlength{\tabcolsep}{#1}\renewcommand{\arraystretch}{#2}\centering\footnotesize}

\begin{document}

\title{OCSampler: Compressing Videos to One Clip with Single-step Sampling}




\author{Jintao Lin$^{1}$ \ \ \ \ \ \ \ \ \ \ \ Haodong Duan$^{2}$ \ \ \ \ \ \ \ \ \ \ \ Kai Chen$^{3,4}$\ \ \ \ \ \ \ \ \ \ \ Dahua Lin$^{2}$\ \ \ \ \ \ \ \ \ \ \ Limin Wang\textsuperscript{$1$ \Letter}\\
$^{1}$ State Key Laboratory for Novel Software Technology, Nanjing University, China\\
$^{2}$ The Chinese University of Hong Kong \ \ $^{3}$ SenseTime Research \ \ $^{4}$ Shanghai AI Laboratory\\
\tt\small jintaolin@smail.nju.edu.cn\ \ dh019@ie.cuhk.edu.hk\ \ chenkai@sensetime.com\\
\tt\small dhlin@ie.cuhk.edu.hk\ \ 07wanglimin@gmail.com
}

\maketitle

\begin{abstract}


In this paper, we propose a framework named OCSampler to explore a compact yet effective video representation with one short clip for efficient video recognition. 
Recent works prefer to formulate frame sampling as a sequential decision task by selecting frames one by one according to their importance, 
while we present a new paradigm of learning instance-specific video condensation policies to select informative frames for representing the entire video only in a single step. 
Our basic motivation is that the efficient video recognition task lies in processing a whole sequence at once rather than picking up frames sequentially. 
Accordingly, these policies are derived from a light-weighted skim network together with a simple yet effective policy network within one step. 
Moreover, we extend the proposed method with a frame number budget, enabling the framework to produce correct predictions in high conﬁdence with as few frames as possible. 
Experiments on four benchmarks, \textit{i.e.}, ActivityNet, Mini-Kinetics, FCVID, Mini-Sports1M, demonstrate the effectiveness of our OCSampler over previous methods in terms of accuracy, theoretical computational expense, actual inference speed. 
We also evaluate its generalization power across different classifiers, sampled frames, and search spaces. 
Especially, we achieve 76.9\% mAP and 21.7 GFLOPs on ActivityNet with an impressive throughput: 123.9 Video/s on a single TITAN Xp GPU.


\end{abstract}
\blfootnote{\Letter: Corresponding author.}

\section{Introduction}

With the explosive popularity of social media platforms as well as bountiful online video content, there comes wider attention on effective and scalable approaches that can deal with actions or events recognition in the face of the video data deluge. To this end, most efforts have been devoted to exploring a sophisticated temporal module to capture relationships across the time dimension by densely applying 2D-CNNs~\cite{lin2019tsm, wang2016temporal, li2020tea, simonyan2014two, feichtenhofer2016convolutional, yue2015beyond} or 3D-CNNs~\cite{tran2015learning, carreira2017quo, feichtenhofer2019slowfast, tran2019video, qiu2017learning}. Although these models achieve superior accuracy performance, their computational expense limits their application in real-world scenarios where the model deployment is resource-constrained and requires to process high data volumes with stringent latency and throughput requirements.

\begin{figure}[t]
    \begin{center}
    \centerline{\includegraphics[width=0.935\columnwidth]{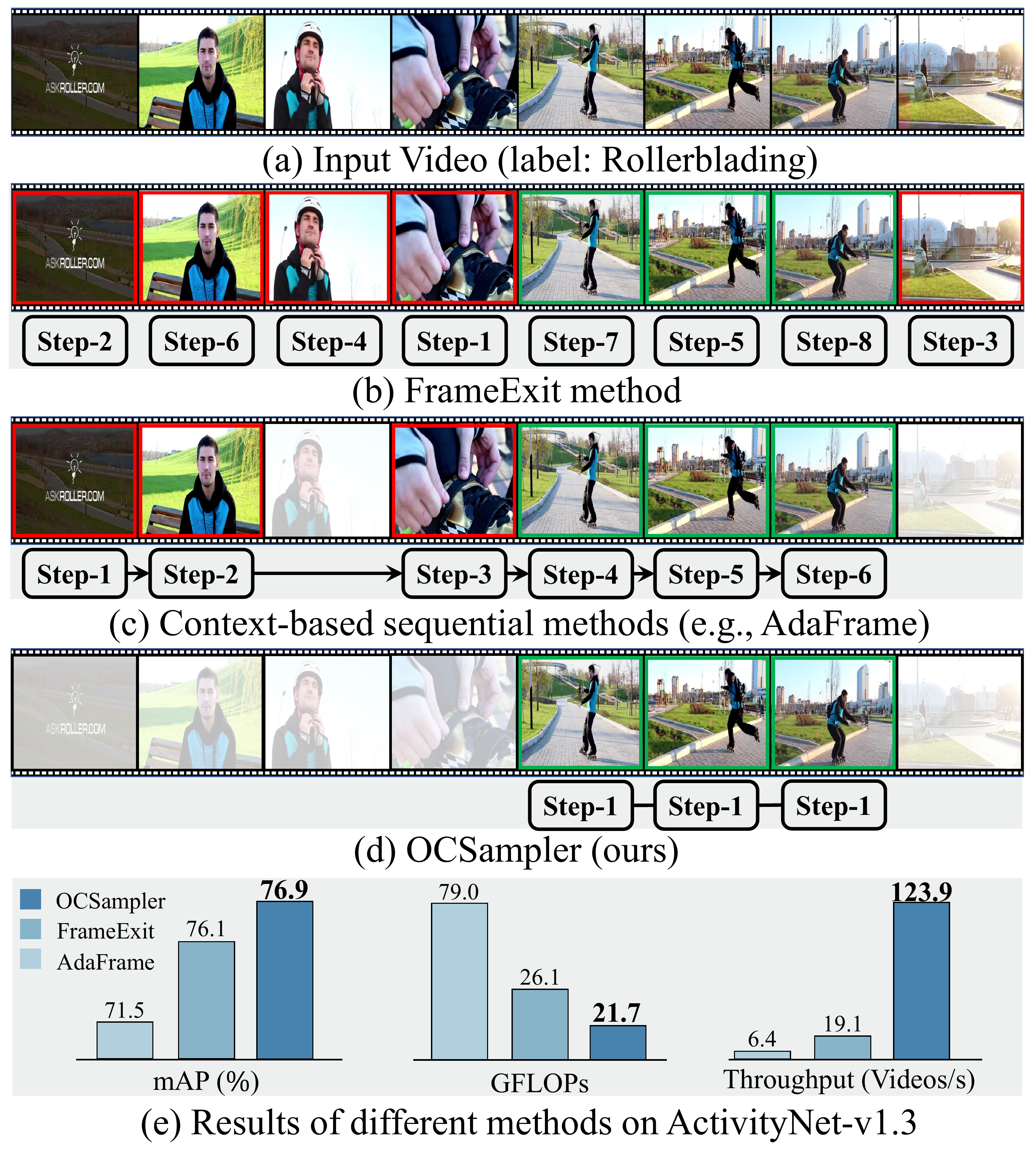}}
    \caption{\textbf{Comparisons of other methods and our proposed OCSampler.} Most existing works reduce computational cost by regarding the frame selection problem as a sequential decision task, while OCSampler aims to perform efficient inference by making one-step decision with holistic views. Our method achieves excellent performance on accuracy, theoretical computational expense, and actual inference throughout.
    }
    \label{fig:inrto}
    \end{center}
    \vspace{-8ex}
\end{figure}

To mitigate this issue, a large body of research has been focusing on designing light-weighted modules~\cite{feichtenhofer2020x3d, tran2018closer, piergiovanni2019tiny, tran2018closer, qiu2017learning, xie2017rethinking, zolfaghari2018eco, karpathy2014large, lin2019tsm} to bring efficiency improvements. Being unaware of the complexity of video contents and instance-specific difficulties for video recognition, these models treat all the videos equally and adopt naive sampling strategies. To overcome this limitation, extensive studies~\cite{ghodrati2021frameexit, wu2019adaframe, gao2020listen, yeung2016end, fan18efficientvideoclassification} have been conducted to devise adaptive mechanisms of frame selection on a per-video basis by either determining which frame to observe next, or conditional early exiting in a deterministic order. These approaches all model the frame selection problem as a sequential decision task and prefer these decisions to be made individually per frame, leaving out the subsequent parts of the video. Thus, these methods require more inference time even with theoretical computational efficiency and lead to sub-optimal results. Recent methods~\cite{meng2020adafuse, meng2020ar, Wang_2021_ICCV, sun2021dynamic, wu2019liteeval, li20202d} rely on designing different preset transformations (\textit{e.g.}, process at a specific spatial resolution~\cite{meng2020ar}, process at a specific patch~\cite{Wang_2021_ICCV}, \textit{etc.}) and determining which action should be taken on each frame or network module to alleviate computational burden. However, the key to video recognition is aggregating features across different frames. Most of these methods rely on the assumption that several salient frames are equally important to an effective video representation for video recognition, which may introduce temporal redundancy and lack specific consideration for temporal modeling.

A promising alternative direction to reduce the computational complexity of analyzing video content, without sacrifice of recognition accuracy, is representing videos with one clip in a single step. Clip-level features~\cite{korbar2019scsampler, tran2015learning, carreira2017quo, feichtenhofer2019slowfast, tran2019video} commonly used in 3D-CNNs methods reveal the superiority owing to its spatio-temporal information extraction. However, traditional clip-level sampling requires to average the predictions of multiple clips, and clips containing visual redundancy will pollute the final results. Inspired by that, we design an efficient video recognition framework that compresses trimmed/untrimmed videos into a single clip by evaluating a clip-based reward on a per-video basis in one turn. As shown in Figure~\ref{fig:inrto}, our basic idea is that modeling the selection problem as a one-step decision task can yield significant savings in both theoretic computation and actual inference budget, and sampling based on an integrated clip is more reasonable than evaluating several frames individually.

In particular, in this paper, we propose a novel OCSampler approach to dynamically localize and attend to the instance-specific condensed clip of each video. More specifically, our method first takes a quick skim over the whole video with a light-weighted CNN to obtain coarse global information. Then we train a simple yet effective policy network on its basis to select the most valuable combination of the clip for the subsequent recognition. This module is learnt with the reinforcement learning algorithm due to its non-differentiability. Finally, we activate a high-capacity classifier to process the selected clip. Inference on clips constructed with a small number of frames, considerable computation overhead can be saved. Our method allocates computation unevenly across the temporal locations of videos according to their contributions to the recognition task, leading to a significant improvement in efficiency yet still with preserved accuracy.

The vanilla OCSampler framework processes all videos using the same number of frames, while the only difference lies in the temporal locations of the selected frames. We show that our method can be extended via adopting an adaptive frame number budget to reduce the computation spent on ``easy" videos. This is achieved by introducing an additional budget network that estimates how many frames should be used for a video, which is optimized by pseudo-labels in a self-supervision way. The algorithm is referred to as OCSampler+.

We evaluate the effectiveness of OCSampler on four efficient video recognition benchmarks, namely ActivityNet~\cite{caba2015activitynet}, Mini-Kinetics~\cite{kay2017kinetics}, FCVID~\cite{jiang2017exploiting}, Mini-Sports1M~\cite{karpathy2014large}. Experimental results show that OCSampler consistently outperforms all the state-of-the-art by large margins in terms of accuracy and efficiency. Especially, we achieve 76.9\% mAP and 21.7 GFLOPs on ActivityNet with an impressive throughput: 123.9 Video/s on a single TITAN Xp GPU. We also demonstrate that the frames sampled by our method can be generalized to boost the efficacy and efficiency of an arbitrary classifier.

\section{Related Work}

\begin{figure*}[htb!]
    \centering
    \includegraphics[width=1.0\linewidth]{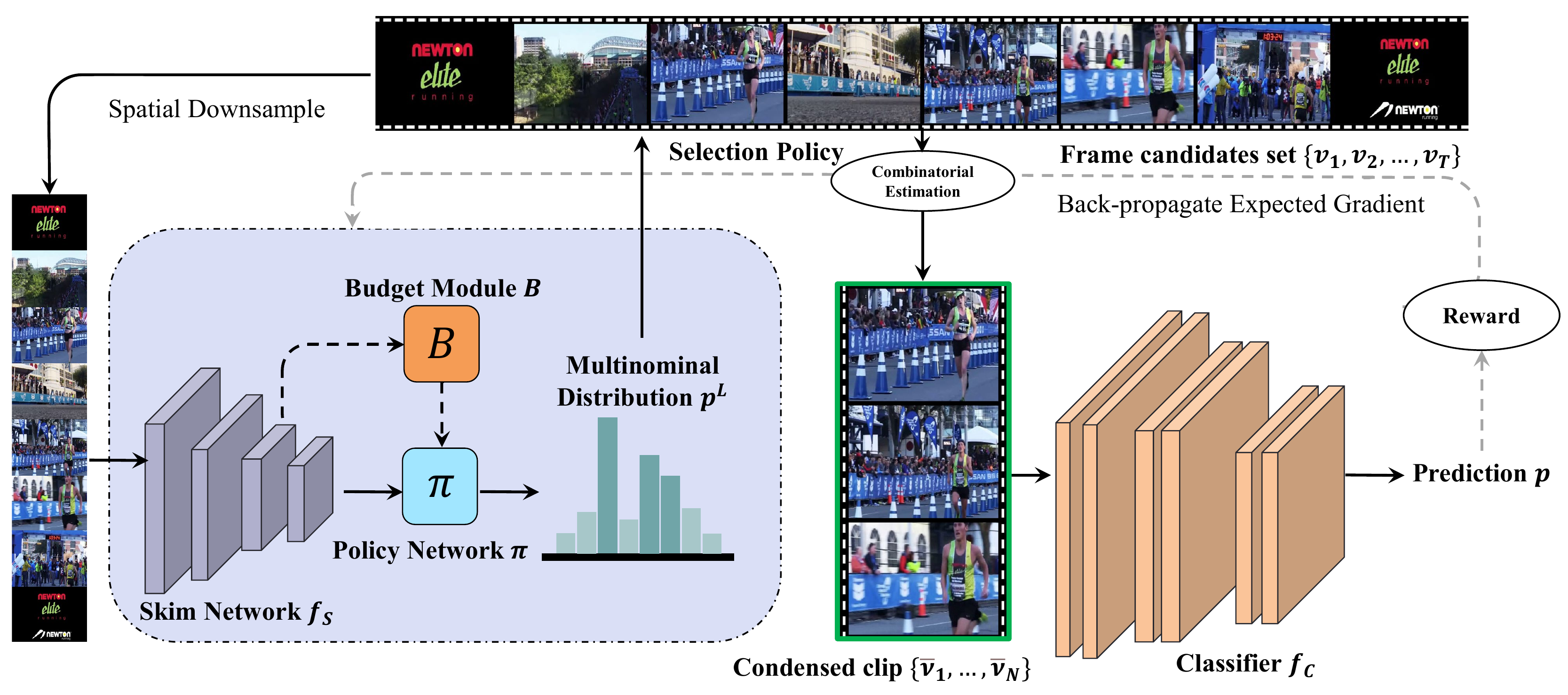}
    \vspace{-0.45cm}
    \caption{\textbf{The overview of our approach.} Given a video, our framework sparsely samples $T$ candidate frames and feeds them into the skim network $f_{S}$ to take a quick look through the video and extract spatio-temporal features. Then a simple policy network is followed to derive a frame selection policy based on the output multi-nominal distribution of $p^L$, which activates a subset of $N$ frames to form a single clip as the product of video condensation. By involving an additional budget module $B$ to determine how many frames should be taken on each video, we can further reduce the redundant computation spent on less important frames. Afterwards, an arbitrary classifier is used to obtain the recognition result. Conditioned on the prediction, we back-propagate the expected gradient with the reward of the integrated clip and the corresponding combinational estimation. See texts for more details.}
    \label{fig:pipeline}
\end{figure*}

\noindent \textbf{Video recognition.}
In the context of deep neural networks, there exist two families of models for video recognition, namely 2D-CNN approaches and 3D-CNN approaches. For 2D-CNN approaches, they commonly equip the state-of-the-art 2D-CNN models with the capability of temporal modeling to aggregate features along the temporal dimension, such as temporal pooling~\cite{wang2016temporal, simonyan2014two, feichtenhofer2016convolutional}, recurrent networks~\cite{donahue2015long, yue2015beyond, li2018videolstm}, efficient temporal modules~\cite{TEINet,lin2019tsm,li2020tea,tam}, and exploiting explicit temporal information like optical flow~\cite{ feichtenhofer2016convolutional, simonyan2014two}. For 3D-CNN approaches~\cite{tran2015learning}, most of the works learn spatial and temporal representation by adopting 3D convolution on stacked adjacent frames. Some of them~\cite{tran2018closer, qiu2017learning} also decompose 3D convolution into a 2D spatial convolution and a 1D temporal convolution or integrate 2D CNN into 3D CNN~\cite{Zhou_2018_CVPR}. However, existing sampling strategies applied to 2D-CNN approaches and 3D-CNN approaches have some shortcomings. Frames uniformly sampled along temporal dimension are sent to 2D-CNN models, which takes fewer frames to represent the whole video but may miss the key information when actions occur in a moment. 3D-CNN models need to aggregate predictions of multiple clips to get a reasonably good result, consuming vast amounts of computation (especially for untrimmed videos). In contrast, our idea is to exploit an effective way to condense a video using a single short clip, which is agnostic to different models.

\noindent \textbf{Sequential sampling.} 
To reduce the theoretical computation costs of video recognition, these approaches consider the frame selection problem as a sequential decision task and require to wait for previous information to indicate which frame to observe next or whether to exit the selection procedure. AdaFrame~\cite{wu2019adaframe} proposed a Memory-augmented LSTM that provides context information for searching which one to observe next over time. ListenToLook~\cite{gao2020listen} proposed to estimate clip information with a single frame and its accompanying audio using a distillation framework. However, using audio as preview information to seek the next frame cannot avoid irrelevant frames and still takes more than one step to get the final prediction result of the entire video. FrameExit~\cite{ghodrati2021frameexit} formulated the problem in an early-exiting framework with a simple sampling strategy. For each video, FrameExit followed a preset policy function to check each coming frame sequentially and threw out an exiting signal to quit the procedure. Although this simple policy function avoids complex calculations, its deterministic sampling pattern is sub-optimal in terms of exploitation and exploration. In practice, these sequential sampling methods~\cite{ghodrati2021frameexit, gao2020listen, wu2019adaframe, fan18efficientvideoclassification, yeung2016end} still consume plenty of inference time due to their complex decision process.

\noindent \textbf{Parallel sampling.} 
To mitigate the above issues, some works adopt parallel sampling, which usually chooses what action should be taken on each frame/clip independently and obtains the final selection in parallel. SCSampler~\cite{korbar2019scsampler} used a light-weighted network to estimate a saliency score for each fixed-length clip, while DSN~\cite{zheng2020dynamic} advanced TSN~\cite{wang2016temporal} framework by dynamically sampling a discriminative frame within each segment. They both performed the sampling procedure in a non-sequential manner at the cost of limited decision space, leading to sub-optimal selection due to the holistic information vacancy. MARL~\cite{wu2019multi} utilized multi-agents to learn to pick important frames in parallel and had to go through a heavy CNN in many iterations to yield STOP actions for all agents. Other works reduced computational overhead by selecting input resolution~\cite{meng2020ar}, choosing image patches~\cite{Wang_2021_ICCV}, or assigning different bits~\cite{sun2021dynamic}. 

In contrast, our method relies on a simple one-step reinforcement learning optimization and does not require multiple steps to determine the final frame selection. Besides, we do not use any RNN-based module but directly aggregate a more holistic feature for video-level modeling. We formulate the problem in a video-to-one-clip condensation framework and show that a reasonable reward function, together with an adaptive frame number budget, can lead to significant performance in both theory and practice.

\section{Method}
\label{sec:method}


Unlike most existing works aiming at promoting efficient video recognition 
by selecting a few frames or clips progressively, 
our goal is to compress a trimmed/untrimmed video into one single clip with as few frames as possible,
while preserving sufficient spatio-temporal cues for video recognition.
To this end, we introduce OCSampler,
an efficient and effective framework to condense a video into an integrated clip.
With OCSampler, the computation overhead can be significantly reduced without sacrificing accuracy.
We first describe the components of OCSampler. 
Then we introduce the training algorithm for each component. 
Finally, we extend our framework by considering an adaptive frame number budget, 
which allocates different amounts of computation for each video. 

\subsection{Network Architecture}
\label{sec:arch}


\noindent \textbf{Overview.}
Figure~\ref{fig:pipeline} illustrates an overview of our approach. 
Given an input video, we first uniformly sample $T$ frames along the temporal dimension as frame candidates. 
OCSampler first skims the frame candidates at a lower resolution using a light-weighted skim network  $f_{\textnormal{S}}$, 
to obtain coarse frame-level features. 
Then, the features are fed into the policy network $\pi$ to encode spatio-temporal information 
across frames and determine the optimal frame set to form an integrated clip, 
which maximizes a reward function parameterized by the output from the classifier $f_{\textnormal{C}}$. 
The classifier $f_{\textnormal{C}}$ takes the single clip as inputs and predicts the action category. 
It is worth noting that OCSampler obtains an integrated clip only in one step. 
In the following sections, we describe these components in details.

\noindent \textbf{Skim network $f_{\textnormal{S}}$ } is a light-weighted network to extract deep features for frame candidates. 
It is designed to provide global views across different time in a video for determining which frames should be selected to form a clip for classifier $f_{\textnormal{C}}$. 
Components like TSM~\cite{lin2019tsm} can be inserted to equip Skim network with the capability of fusing information among frame candidates. Note that the additional computation cost incurred by $f_{\textnormal{S}}$ is negligible compared with the classifier $f_{\textnormal{C}}$. 

Formally, given a frame candidate set $\{\bm{v}_1, \bm{v}_2, \ldots, \bm{v}_T\}$ uniformly sampled along the temporal dimension in a video with spatial size $H\!\times\!W$, they are first resized to lower resolution $\tilde{H}\!\times\!\tilde{W}$ and then sent to $f_{\textnormal{S}}$ to generate a global video descriptor $\bm{z}^{\textnormal{S}}_{t}$:
\begin{equation}
    \bm{z}^{\textnormal{S}} = \{\bm{z}^{\textnormal{S}}_{1}, \bm{z}^{\textnormal{S}}_{2}, \ldots, \bm{z}^{\textnormal{S}}_{T}\} = f_{\textnormal{S}}(\{\tilde{\bm{v}}_1, \tilde{\bm{v}}_2, \ldots, \tilde{\bm{v}}_T\}),
\end{equation}
where $t$ is the frame index and $\bm{z}^{\textnormal{S}}_{t}$ encodes context information for each frame on a per-video basis. 

\begin{figure}[t]
    \begin{center}
    \centerline{\includegraphics[width=0.98\columnwidth]{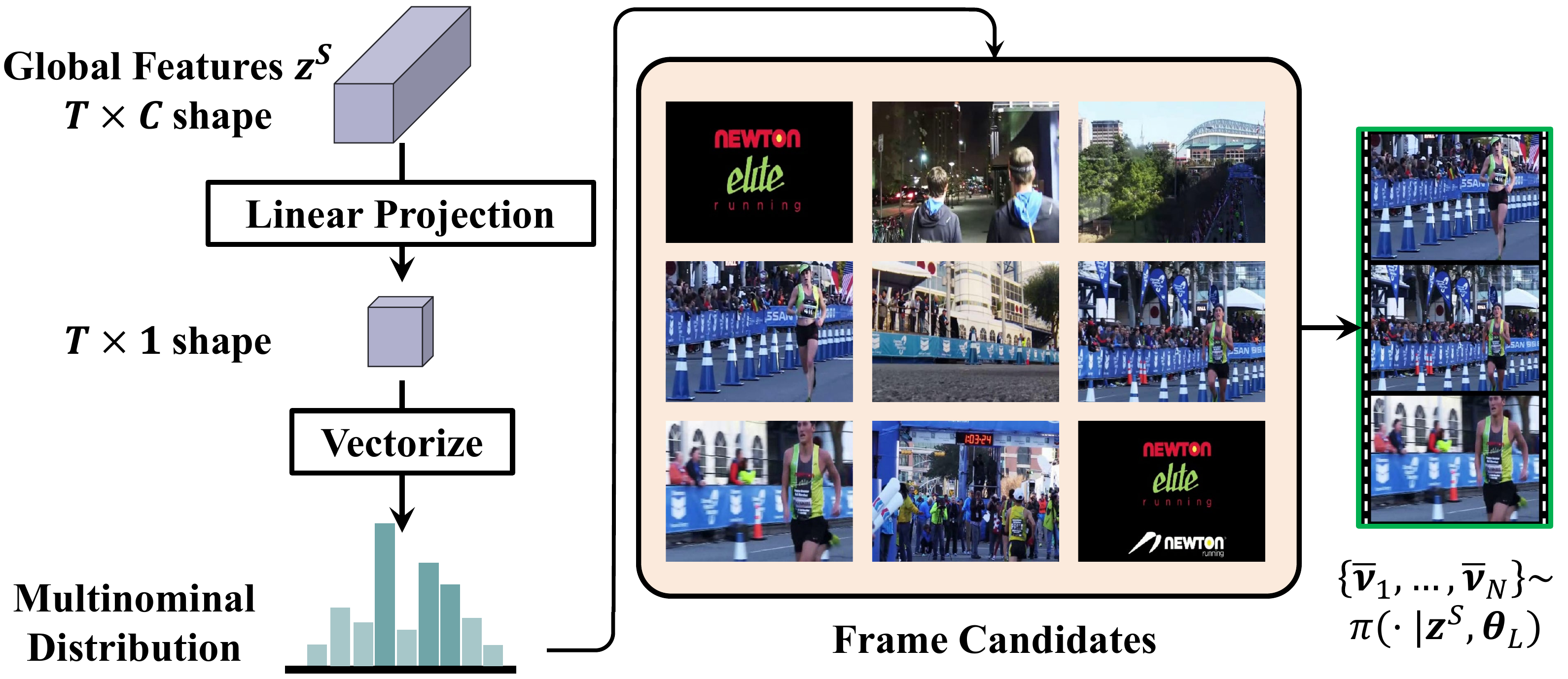}}
    \vspace{-0.2cm}
    \caption{\textbf{The architecture of the policy network.} The global context feature $z^{S}$ is fed into a linear projection layer followed by a vectorization operation, the output of which establish a multinomial distribution $\pi(\cdot|\bm{z}^{\textnormal{S}}, \theta_L)$ on frame candidates (here we take 9 as an example). During training, we sample frames $\overline{\bm{v}}_1, \overline{\bm{v}}_2, \ldots, \overline{\bm{v}}_N\}$ from $\pi(\cdot|\bm{z}^{\textnormal{S}}, \theta_L)$, while at the test time, we directly select frames with the largest $N$ softmax probability.}
    \label{fig:policy_img}
    \end{center}
    \vspace{-4.5ex}
\end{figure}

\noindent \textbf{Policy network $\pi$} receives the global context feature $\bm{z}^{\textnormal{S}}$ from Skim network $f_{\textnormal{S}}$, and localizes which frames can be used to form a salient clip for each video. Note that this procedure is performed only in one iteration and uses no complicated CNN-based or RNN-based modules but one linear projection $f_{\textnormal{L}}$ followed by Softmax function $\phi$ with an effective clip-relevant policy function:
\begin{equation}
    \label{eq:predict_prob}
    \bm{p}^{\textnormal{L}} = \{p^{\textnormal{L}}_1, p^{\textnormal{L}}_2, \ldots, p^{\textnormal{L}}_{T}\} = \phi(f_{\textnormal{L}}(\{\bm{z}^{\textnormal{S}}_{1}, \bm{z}^{\textnormal{S}}_{2}, \ldots, \bm{z}^{\textnormal{S}}_{T}\})),
\end{equation}
where $\bm{p}^{\textnormal{L}}_t$ refers to the softmax probability for each frame. Formally, as shown in Figure~\ref{fig:policy_img}, $\pi$ determines the chosen $N$ frames from candidates $\{\bm{v}_1, \bm{v}_2, \ldots, \bm{v}_T\}$ to be sent to classifier $f_{\textnormal{C}}$. Since the target is to determine a representative clip rather than several salient frames, it involves making set-level decisions that are non-differentiable and harder than making binary ones due to larger search space. Given that, we still formalize $\pi$ as a one-step Markov Decision Process (MDP) and train it with reinforcement learning. Specifically, the selection of the clip $\{\overline{\bm{v}}_1, \overline{\bm{v}}_2, \ldots, \overline{\bm{v}}_N\}$ is drawn from the distribution $\pi(\cdot|\bm{z}^{\textnormal{S}}, \theta_L)$.


where $\theta_{L}$ denotes learnable parameters of the linear projection $f_{\textnormal{L}}$. In our implementation, we establish a multinomial distribution on them, parameterized by the output probability of $\pi$. During training, $\{\overline{\bm{v}}_1, \overline{\bm{v}}_2, \ldots, \overline{\bm{v}}_N\}$ are produced by sampling from the policy based on corresponding multinomial distribution. During testing, candidates with maximum probabilities are adopted in a deterministic inference procedure. 

\noindent \textbf{Classifier $f_{\textnormal{C}}$} can be any classification network used in video recognition. It receives a clip of temporal length $N$ from policy network $\pi$ and outputs the recognition result of the video. To be specific, Classifier $f_{\textnormal{C}}$ directly processes a clip of $N$ frames $\{\overline{\bm{v}}_1, \overline{\bm{v}}_2, \ldots, \overline{\bm{v}}_N\}$ with original resolution $H\!\times\!W$, \textit{i.e.},
\begin{equation}
    \label{eq:classifier}
    \bm{p} = f_{\textnormal{C}}(\{\overline{\bm{v}}_1, \overline{\bm{v}}_2, \ldots, \overline{\bm{v}}_N\}),
\end{equation}
where $\bm{p}$ indicates the probability scores for each class. Notably, Classifier $f_{\textnormal{C}}$ accounts for most of the computational overhead in our framework and yields the prediction at a time, instead of sequentially processing each frame. Such a design reduces both computational complexity in theory and inference time in practice.

\subsection{Training Algorithm}
\label{sec:training}

There are two stages in our training algorithm to optimize OCSampler framework.

\noindent \textbf{Stage I: Initialization.} In this stage, we warm up $f_{\textnormal{S}}$ and $f_{\textnormal{C}}$ by video recognition tasks on target datasets. In specific, we train $f_{\textnormal{S}}$ by randomly sampling $T$ frames with size $\tilde{H}\!\times\!\tilde{W}$ to minimize the cross-entropy loss $L_{\textnormal{CE}}(\cdot)$ over the training set $\mathcal{D}_{\textnormal{train}}$:
\begin{equation}
    \label{eq:stage_1_S}
    \begin{split}
        \mathop{\textnormal{minimize}}_{f_{\textnormal{S}}}\ \ \  \mathbb{E}&_{\{\tilde{\bm{v}}_1, \tilde{\bm{v}}_2, \ldots, \tilde{\bm{v}}_T\} \in \mathcal{D}_{\textnormal{train}}}
    \left[L_{\textnormal{CE}}(\tilde{\bm{p}}, y)\right].
    \end{split}
\end{equation}
Similarly, we pretrain $f_{\textnormal{C}}$ by using randomly sampled $N$ frames with $H\!\times\!W$ resolution:
\begin{equation}
    \label{eq:stage_1_C}
    \begin{split}
        \mathop{\textnormal{minimize}}_{f_{\textnormal{C}}}\ \ \  \mathbb{E}&_{\{\bm{v}_1, \bm{v}_2, \ldots, \bm{v}_N\} \in \mathcal{D}_{\textnormal{train}}}
    \left[L_{\textnormal{CE}}(\bm{p}, y)\right].
    \end{split}
\end{equation}
Here, $y$ refers to the corresponding label of the sample. Given the good recognition performance,  $f_{\textnormal{S}}$ and $f_{\textnormal{C}}$ are equipped with the ability to extract spatio-temporal features from an arbitrary sample on target datasets and provide good quality reward signals with less noise, leaving the basis for policy network $\pi$.

\noindent \textbf{Stage II: Optimizing policy network.} In this stage, we freeze the parameters of classifier $f_{\textnormal{C}}$ learned in stage I and train policy network $\pi$ with reinforcement learning by solving one-step Markov Decision Process problem. Based on the probability $\bm{p}^{\textnormal{L}}$ predicted by $f_{\textnormal{L}}$ with global context feature $\bm{z}^{\textnormal{S}}$ (see Eq.~\ref{eq:predict_prob}), $\pi$ receives a reward $r$ indicating how beneficial this combination is to construct a clip for recognition. We optimize $\pi$ by maximizing the sum of the rewards:
\begin{equation}
    \label{eq:stage_2_all}
    \mathop{\textnormal{maximize}}_{\pi}\ \ \  \mathbb{E}_{\{\overline{\bm{v}}_1, \overline{\bm{v}}_2, \ldots, \overline{\bm{v}}_N\}\sim\pi(\cdot|\bm{z}^{\textnormal{S}}, \theta_L)}
    \left[
        r
    \right].
\end{equation}
In our implementation, we adopt the off-the-shelf policy gradient algorithm~\cite{williams1992simple} to solve Eq.~\ref{eq:stage_2_all}. Note that there are $\tbinom{T}{N}$ different cases to choose $N$ frames from $T$ candidates, which makes it hard to precisely calculate the combinatorial probability and intractable to handle directly. Formally, we define $q(i_1, \ldots, i_{N} | \bm{p}^{\textnormal{L}})$ as the probability of sampling frames sequentially with the order $(i_1, \ldots, i_{N})$:
\begin{equation}
    \label{eq:sequence}
    \small
    q(i_1, \ldots, i_{N}|\bm{p}^{\textnormal{L}})=p^{L}_{i_1}\!\times\!\frac{p^{L}_{i_2}}{1-p^{L}_{i_1}}\!\times\ldots\!\times\frac{p^{L}_{i_N}}{1-\sum_{j=1}^{N-1}p^{L}_{i_j}},
\end{equation}
There are $N!$ different permutations for $N$ elements, we denote the set of all $N!$ as $\cP$. Then the probability of sampling these $N$ frames can be precisely calculated by summing $q$ for all $N!$ different permutations:
\begin{equation}
    \label{eq:sumprob}
    \small
    Prob_{\{\overline{\bm{v}}_1, \overline{\bm{v}}_2, \ldots, \overline{\bm{v}}_N\}}=\sum_{\sigma \in {\cP}} q(\sigma(i_1), \sigma(i_2), ..., \sigma(i_N)|\bm{p}^{\textnormal{L}}).
\end{equation}
However, Eq.~\ref{eq:sumprob} is only tractable for a small $N$ (\textit{e.g.}, $N\!<\!10$). In experiments, we estimate this term with the probability of a subset of all permutations (\textit{e.g.}, subset with $\tbinom{T}{8}$ items) and find that the policy network can be optimized well either with the precise or the estimated probability. 

In our case, where policy network aims at figuring out how to condense a video with one clip rather than pick up several frames separately, the reward $r$ is expected to evaluate the integrated clip $\overline{V}$, \textit{i.e.}, $\{\overline{\bm{v}}_1, \overline{\bm{v}}_2, \ldots, \overline{\bm{v}}_N\}$, in terms of video recognition. To this end, we define $r$ as:
\begin{equation}
    \label{eq:reward}
    \begin{split}
        &r(\{\overline{\bm{v}}_1, \ldots, \overline{\bm{v}}_N\}) \\=\  &\bm{p}_{y}(\{\overline{\bm{v}}_1, \ldots, \overline{\bm{v}}_N\}) \\  &- \ {\mathbb{E}}_{\overline{V}\sim\textnormal{UniformSample}(\{\bm{v}_1, \ldots, \bm{v}_T\})}\left[\bm{p}_{y}(\overline{V})\right],
    \end{split}
\end{equation}
where $\bm{p}_{y}$ refers to the softmax prediction on $y$ (\textit{i.e.}, confidence on the ground-truth label, see Eq.~\ref{eq:classifier}). When computing $r$, we take all of the $N$ frames $\{\overline{\bm{v}}_1, \ldots, \overline{\bm{v}}_N\}$ into consideration to avoid information redundancy and short-sighted mistakes raised by single frame judgement. The second term in Eq.~\ref{eq:reward} refers to the expected value obtained by uniformly sampling $N$ frames from candidates. Since reinforcement learning may be of high variance and converge slowly, we introduce another policy, which does not depend on the policy network, to affect the variance and stabilize the training process significantly. 

\subsection{Adaptive Frame Number Budget}
\label{sec:ada_number}

Processing videos of different complexity equivalently with the same amount of computation is still sub-optimal.
To overcome this, we extend our OCSampler to OCSampler+, which automatically learns to select fewer frames for easier videos and more frames for harder ones.

\noindent {\bf Budget module.} We add an additional Budget module $f_{\textnormal{B}}$ that takes global context feature $\bm{z}^{\textnormal{S}}$ as input and is inserted between Skim network $f_{\textnormal{S}}$ and policy network $\pi$. Each of these features is first passed to one layer of MLP with 64 neurons independently (shared weights among all streams). The resulting features are then averaged and linearly projected, followed by a softmax function to estimate the frame number budgets.

\begin{figure}[t]
    \centering
    \resizebox{\columnwidth}{!}{\includegraphics[width=1\linewidth]{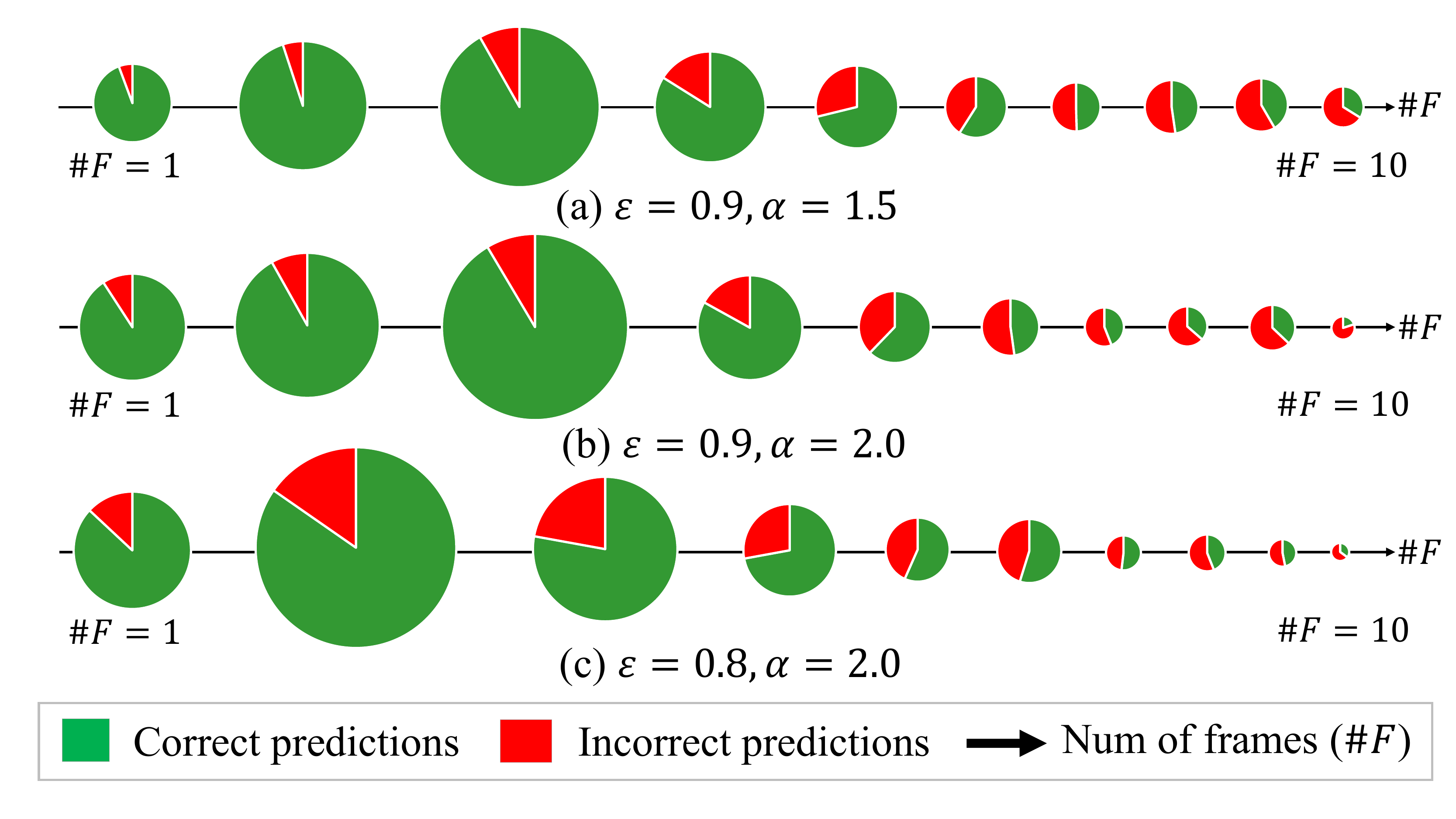}}
    \vspace{-0.8cm}
    \caption{\textbf{Trade-off between frame number budgets and prediction accuracy.} The statistics of our method equipped with a budget module for different $\epsilon$ and $\alpha$ on the validation set of ActivityNet. The circle area at a certain number of $\#F$ represents the percentage of samples using $\#F$ frames for prediction. Easier examples use fewer frames with higher accuracy, while harder examples use more frames leading to increased miss-classifications. }
    \vspace{-0.3cm}
    \label{fig:watermelon}
\end{figure}

\noindent {\bf Training with Self-Supervision.} We construct a budget label $y^B$ indicating the probability of how many frames should be used by analyzing the statistics obtained from considering all of the combinations.
Formally, given a video, we define $\cG^{m}\!=\!\{g^{m}_1, g^{m}_2, \ldots, g^{m}_{c}\}$ (where $1\!\leq\!{m}\!\leq{T}$ and $c\!=\!{\tbinom{T}{m}}$) as the list containing combinations of $m$ frames from the frame candidate set $\{\bm{v}_1, \bm{v}_2, \ldots, \bm{v}_T\}$. We send each item $g^{m}_{i} \in \cG^{m}$ to classifier $f_{\textnormal{C}}$ to obtain a boolean value $a^{m}_{i}\!\in\!\{0,1\}$ 
, which specifies whether this combination can be predicted correctly. 
After that, we obtain the ratio of prediction correction $r^m$ with the estimation:
\begin{equation}
    \label{eq:p_combin}
    r^{m}={\sum_i a_i^m}/{\tbinom{T}{m}}.
\end{equation}
Based on $r^m$, we use $\epsilon$ to determine the minimum budget required to predict a video correctly with classifier $f_{\textnormal{C}}$:
\begin{equation}
    \label{eq:argmin}
    y^{B}_{k}=1,\ \ \mathrm{where} \ k=\mathop{\arg\min}_{i}\ (\epsilon\!\leq\!{r^{i}}).
\end{equation}
Provided that single-label is more likely to lead to bias on accuracy, we leverage other options with a smooth function to balance the accuracy and efficiency:
\begin{equation}
    \label{eq:budget_label}
    y^{B}_{i} = 
    \begin{cases}
      0 & \text{if $i<k$}, \\
      \frac{1}{\alpha^{(i-k)}} & \text{if $i>k$} ,\\
    \end{cases}
\end{equation}
where $\alpha\textgreater1$ and is the hyper-parameter that controls the trade-off between model accuracy and computational cost. An example is shown in Figure~\ref{fig:watermelon}. Then, we learn the parameters of the budget network by minimizing the cross-entropy loss between the predicted probability and the pseudo label $y^{B}$:
\begin{equation}
    \label{eq:L_Budget}
    L_{\textnormal{Budget}}=L_{\textnormal{CE}}(\bm{z}^{\textnormal{S}}, y^{B}).
\end{equation}

Notably, this procedure of estimating frame budgets also applies for one step. Similar to Eq.~\ref{eq:sumprob}, we use Monte-Carlo sampling to estimate $r^m$ for Eq.~\ref{eq:p_combin}. Moreover, to overcome the long-tail issue owing to sample imbalance, we assign class weight based on the sample distribution for Eq.~\ref{eq:L_Budget}. During training, we first optimize the Budget module $f_{\textnormal{B}}$ with skim network $f_{\textnormal{S}}$ to get the frame budget estimation, and then learn the policy network $\pi$ as mentioned in Stage II. During inference, we choose the maximum probability in $f_{\textnormal{B}}$ as the number of used frames.

\section{Experiment}
\label{sec:exp}



\begin{figure}[t]
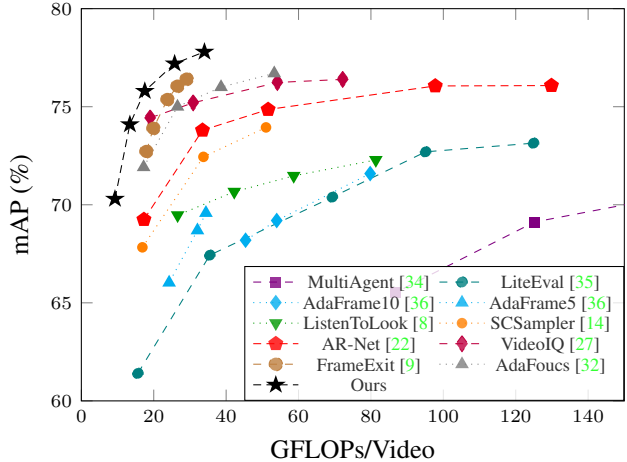

\noindent
\centering
\pgfplotsset{
tick label style={font=\scriptsize},
xlabel style={font=\scriptsize},
ylabel style={font=\scriptsize},
legend style={font=\scriptsize, at={(1,0)},anchor=south east, fill opacity=0.8},
compat=newest,
}
\begin{tikzpicture}[trim left=-0.4cm]
    \begin{axis}[
     legend columns=2,
      height=6.8cm,
	  width=8.8cm,
      xlabel=GFLOPs/Video,
      ylabel=mAP (\%),
      xmin=0,
      xmax=150,
      ymin=60,
      ymax=80,
      legend style={nodes={inner sep=1.5pt,text depth=0.0em}}],
      xtick={0,25,50,75,100,125,150},
      ytick={60,62,64,66,68,70,72,74,76,78,80},
    ]
    \input{figures/my_tikz/data/data_multiagent}
    \input{figures/my_tikz/data/data_liteeval}
    \input{figures/my_tikz/data/data_adaframe10}
    \input{figures/my_tikz/data/data_adaframe5}
    \input{figures/my_tikz/data/data_ltl_mobilenet_resnet}
    \input{figures/my_tikz/data/data_scsampler}
    \input{figures/my_tikz/data/data_arnet}
    \input{figures/my_tikz/data/data_videoiq}
    \input{figures/my_tikz/data/data_frameexit}
    \input{figures/my_tikz/data/data_adafocus}
    \input{figures/my_tikz/data/ours}
    \legend{
    MultiAgent \cite{wu2019multi}\\
    LiteEval \cite{wu2019liteeval}\\
    AdaFrame10 \cite{wu2019adaframe}\\
    AdaFrame5 \cite{wu2019adaframe}\\
    ListenToLook \cite{gao2020listen}\\
    SCSampler \cite{korbar2019scsampler}\\
    AR-Net \cite{meng2020ar}\\
    VideoIQ \cite{sun2021dynamic}\\
    FrameExit \cite{ghodrati2021frameexit}\\
    AdaFoucs \cite{Wang_2021_ICCV}\\
    Ours\\
    }
    \end{axis}
\end{tikzpicture}
\vspace{-4mm}
\caption{\textbf{Accuracy vs. efficiency curves on ActivityNet.} Our proposed OCSampler obtains the best recognition accuracy with fewer GFLOPs than state-of-the-art methods. We directly quote the numbers reported in published papers.}
\label{fig:flops_plt}
\vspace{-5mm}
\end{figure}

In this section, we conduct comprehensive experiments on four widely used datasets to verify our proposed method. We first briefly describe our experimental setup. Then, we compare OCSampler with some state-of-the-art approaches for efficient video understanding, showing that OCSampler boosts the performance of existing methods.
Finally, we provide ablation results to provide additional insights into our policy learning.

\subsection{Experimental Setup}
\label{sec:exp:setup}
\noindent \textbf{Datasets.}
\label{sec:exp:datasets}
We report the performance of our approach on four datasets:
(1) ActivityNet-v1.3~\cite{caba2015activitynet} consists of 200 classes and contains 10,024 training videos and 4,926 validation videos with an average duration of 117 seconds;
(2) FCVID~\cite{jiang2017exploiting} is labeled with 239 action categories and includes 45,611 training videos and 45,612 validation videos with an average duration of 167 seconds;
(3) Mini-Kinetics has 200 classes from Kinetics~\cite{kay2017kinetics} assembled by~\cite{meng2020ar, meng2020adafuse}, including 121,215 training videos and 9,867 validation videos with an average duration of 10 seconds;
(4) Mini-Sports1M is a subset of full Sports1M~\cite{karpathy2014large} introduced by~\cite{gao2020listen}, containing 30 training videos per class and 10 validation videos per class with a total of 487 action classes.

\noindent \textbf{Evaluation metrics.}
\label{sec:exp:metrics}
To evaluate the accuracy, We use top-1 accuracy for multi-class (Mini-Kinetics) classification and mean average precision (mAP) for multi-label classification (ActivityNet, FCVID, and Mini-Sports1M), respectively. To measure the computational cost, we use giga floating-point operation (GFLOPs) as efficiency reflection, which is a hardware-independent metric. 
We report per video GFLOPs for all experiments since some methods use different numbers of frames per video for recognition.



\begin{table}[t]
    \centering
    \begin{footnotesize}
    \caption{\textbf{Comparison to state of the art on ActivityNet-v1.3 and Mini-Kinetics.} OCSampler outperforms exiting methods in terms of accuracy and efficiency using ResNet, SlowOnly, and X3D-S backbones with ImageNet pretraining or Kinetics-pretraining. The column of Backbones is for classifier, and best results are \textbf{bold-faced}.}
    \label{table:sota_anet_k200}
    \vspace{-3mm}
    
    \resizebox{\linewidth}{!}{
		\tablestyle{4pt}{1.15}
    \begin{tabular}{cccccc}
    \hline
    \multirow{2}{*}{Methods} & \multirow{2}{*}{Backbones}  & \multicolumn{2}{c}{ActivityNet} &  \multicolumn{2}{c}{Mini-Kinetics} \\
    && mAP & GFLOPs & Top-1 &  GFLOPs \\
    \hline
    \rowcolor{mygray}\textit{ImageNet} & & & & &\\
    LiteEval \cite{wu2019liteeval} & ResNet & \ \ 72.7\% & 95.1 & \ \ 61.0\% & 99.0 \\
    SCSampler \cite{korbar2019scsampler} & ResNet & \ \ 72.9\%  & 42.0 & \ \ 70.8\% & 41.9 \\
    AR-Net \cite{meng2020ar} & ResNet & \ \ 73.8\% & 33.5 & \ \ 71.7\% & 32.0 \\
    videoIQ \cite{sun2021dynamic} & ResNet & \ \ 74.8\% & 28.1 & \ \ 72.3\% & 20.4 \\
    AdaFocus \cite{Wang_2021_ICCV} & ResNet & \ \ 75.0\% & 26.6 & \ \ 72.9\% & 38.6 \\
    FrameExit \cite{ghodrati2021frameexit} & ResNet & \ \ 76.1\% & 26.1 & \ \ 72.8\% & 19.7 \\
    \textbf{OCSampler} & ResNet & \ \ \textbf{77.2\%} & 25.8 & \ \ \textbf{73.7\%} & 21.6 \\
    \textbf{OCSampler} & ResNet & \ \ 76.9\% & \textbf{21.7} & \ \ 72.9\% & \textbf{17.5} \\
    \textbf{OCSampler+} & ResNet & \ \ 75.4\% & 17.9 & \ \ 72.2\% & 15.8 \\
    \hline
    \rowcolor{mygray}\textit{Kinetics} & & & & &\\
    Ada2D \cite{li20202d} & SlowOnly-50 & \ \ 84.0\% & 701 & 79.2\% & 738 \\
    ListenToLook \cite{gao2020listen} & R(2+1)D-152 & \ \ 89.9\% & 2640 & -- & -- \\
    MARL \cite{wu2019multi} & SEResNeXt-152 & \ \ 90.0\% & 7540 & -- & -- \\
    \textbf{OCSampler} & SlowOnly-50 & \ \ 87.3\% & \textbf{68.2} & \textbf{82.6\%} & \textbf{27.3} \\
    \textbf{OCSampler} & SlowOnly-101 & \ \ \textbf{90.1\%} & 593 & - & - \\
    \hline
    \rowcolor{mygray}\textit{Kinetics} & & & & &\\
    FrameExit \cite{ghodrati2021frameexit} & X3D-S & \ \ 86.0\% & 9.8 & \ \ -- & -- \\
    \textbf{OCSampler} & X3D-S & \ \ \textbf{86.6\%} & \textbf{7.9} & \ \ -- & -- \\
    \hline
    \end{tabular}
    }
    \vspace{1mm}
    \end{footnotesize}
\end{table}

\noindent \textbf{Implementation details.}
\label{sec:exp:prepocess}
If not specified, we uniformly sampled 10 frames from each video as frame candidates on all the four datasets. Following~\cite{meng2020ar, ghodrati2021frameexit}, during training, we adopt random scaling to all frames followed by $224 \times 224$ random cropping and random flipping. For the input to light-weighted CNN, we further lower the resolution of video frames to $128 \times 128$. During inference, we still feed light-weighted CNN with $128 \times 128$ resolution frames and average prediction of $224 \times 224$ center-cropped patches for all sampled frames. If not mentioned, we adopt MobileNetV2-TSM and ResNet50 as skim network $f_{\textnormal{S}}$ and classifier $f_{\textnormal{C}}$ respectively. A one-layer fully-connected network with a hidden size of 1280 is used in policy network $\pi$. $T$ is set to 10 by default. 


\subsection{Main Results and Analysis}
\label{sec:exp:main_results}

\noindent \textbf{Comparison with the state-of-the-art methods.}
\label{sec:exp:sota}
The result for ActivityNet and Mini-Kinetics are shown in Table~\ref{table:sota_anet_k200}. For ImageNet-pretrained cases, we use the ResNet-50 model provided by~\cite{ghodrati2021frameexit} as the classifier backbone and use $T=10$ to keep the same with~\cite{ghodrati2021frameexit}. OCSampler outperforms all other approaches by obtaining an enhanced accuracy with up to $5\times$ GFLOPs reduction for both ActivityNet and Mini-Kinetics. Particularly, we outperform all previous methods with more than 4.4 GFLOPs on ActivityNet, and achieve the same Top-1 accuracy with AdaFocus~\cite{Wang_2021_ICCV} using less GFLOPs than half of its on Mini-Kinetics. For Kinetics-pretrained cases, we use SlowOnly models as classifier backbones, and it can be observed that our method outperforms alternative baselines by large margins in terms of efficiency. In particular, on ActivityNet, we outperform MARL~\cite{wu2019multi}, the leading method among competitors, with $11.7\times$ less computational overhead. And for Mini-Kinetics, we also surpass Ada2D~\cite{li20202d} with 3.4\% higher accuracy and $26.0\times$ less GFLOPs. The gain in accuracy is mainly attributed to the larger search space without limitation in our framework, while the gain in efficiency is attributed to the reasonable reward function for video condensation (see Section~\ref{sec:exp:ablation} for detailed analysis).
To verify that the performance of our framework is not limited to the type of classifiers, we conduct experiments with the X3D-S backbone following~\cite{ghodrati2021frameexit}. With the same light-weight X3D-S as our backbone, OCSampler achieves higher accuracy with 1.9\% less GFLOPs, saving 13 frames for inference. This demonstrates the superiority of our framework for efficient video recognition with any classifiers.

\begin{table}[t]
    \centering
    \caption{\textbf{Practical efficiency performance of OCSampler and other currently proposed methods on ActivityNet.} The throughtput are evaluated on a NVIDIA TITAN Xp GPU. Here we use MN, MN-T, RN and SLOW to denote MobileNetV2, MobileNetV2-TSM, ResNet and SlowOnly respectively. The best results are \textbf{bold-faced}.}
    \vspace{-3mm}
    \resizebox{\linewidth}{!}{
		\tablestyle{4pt}{1.15}
    \begin{tabular}{cccccc}
    \hline
    \multirow{2}{*}{Methods} & \multirow{2}{*}{Backbones} & \multirow{2}{*}{mAP} & \multirow{2}{*}{GFLOPs} &  \multicolumn{2}{c}{Throughput}\\
    &&&& (Videos/s) \\
    \hline
    \rowcolor{mygray}\textit{ImageNet} & & & & &\\
    AdaFrame~\cite{wu2019adaframe} & MN+R50 & 71.5\% & 79.0 & 6.4 \\
    FrameExit~\cite{ghodrati2021frameexit} & ResNet-50 & 76.1\% & 26.1 & 19.1 \\
    AR-Net~\cite{meng2020ar} & MN+RN & 73.8\% & 33.4 & 23.1 \\
    AdaFocus~\cite{Wang_2021_ICCV} & MN+RN & 75.0\% & 26.6 & 44.9 \\
    \textbf{OCSampler} & MN-T+R50 & \textbf{76.9\%} & \textbf{21.7} & \textbf{123.9} (\textcolor{blue}{$\uparrow$2.8x}) \\
    \hline
    \rowcolor{mygray}\textit{Kinetics} & & & & &\\
    MARL~\cite{wu2019multi} & SEResNeXt-152 & 90.0\% & 7715 & 0.5 \\
    ListenToLook~\cite{gao2020listen} & (R2+1)D-152 & 89.9\% & 2640 & 0.8 \\
    \textbf{OCSampler} & MN-T+SLOW101 & \textbf{90.1\%} & \textbf{593} & \textbf{4.4} (\textcolor{blue}{$\uparrow$5.5x}) \\
    \hline
    \end{tabular}}
    \vspace{1mm}
    \label{tab:speed}
    \vspace{-5mm}
\end{table}

\noindent \textbf{Results of varying number of used frames} 
are presented in Figure \ref{fig:flops_plt}. We change the number of used frames within $N\!\in$\{2, 3, 4, 6, 8\}, and plot the corresponding mAP v.s. GFLOPs trade-off curves on ActivityNet. We also present current state-of-the-art with various computational costs. One can observe that OCSampler leads to a considerably better trade-off between efficiency and accuracy.

\noindent  \textbf{Adaptive frame number budget.}
We investigate the effectiveness of extended OCSampler with frame number budgets by altering the amount of computational overhead per video. Figure~\ref{fig:watermelon} illustrates accuracy and the number of processed frames with different values of $\alpha$ and $\epsilon$. According to Eq.~\ref{eq:argmin} and Eq.~\ref{eq:budget_label}, a higher $\alpha$ encourages more videos to use fewer frames for recognition (the first row) compared to a lower $\alpha$ (the second row), while a higher $\epsilon$ serves as a more strict threshold to depress using fewer frames for recognition (the second row) compared to a lower $\epsilon$ (the third row). It can also be seen that the fewer number of frames are used, the more correct the result becomes. This trend is desirable since easier samples require less computational cost while harder ones take more overhead.

\noindent  \textbf{Practical efficiency.}
To gain a better understanding of the efficiency achieved by OCSampler, we also test the real inference speed of different methods on a single NVIDIA TITAN Xp GPU. Table~\ref{tab:speed} shows that our practical acceleration is significant compared to other approaches, which is attributed to the one-step decision procedure for all frames without multiple iterations in our framework.


\begin{table}
    \begin{center}
    \caption{\small \textbf{Comparison with state of the art methods on Mini-Sports1M and FCVID}. OCSampler achieves the best mAP while offering significant savings in GFLOPs.}
    \vspace{-2mm}
    \resizebox{\linewidth}{!}{
		\tablestyle{8pt}{1.15}
    \begin{tabular}{ccccc}
        \hline
        \multirow{2}{*}{Methods} &  \multicolumn{2}{c}{Mini-Sports1M} & \multicolumn{2}{c}{FCVID}  \\
        &  mAP & GFLOPs  &  mAP & GFLOPs \\
        \hline 
        LiteEval~\cite{wu2019liteeval} & 44.7\% & 66.2 & 80.0\% & 94.3 \\
        SCSampler~\cite{korbar2019scsampler} & 44.3\% & 42.0 & 81.0\% & 42.0 \\
        AR-Net~\cite{meng2020ar} & 45.0\% & 37.6 & 81.3\% & 35.1 \\
        AdaFuse~\cite{meng2020adafuse} &  44.1\% & 60.3 & 81.6\% & 45.0 \\ %
        \hline 
        \textbf{OCSampler} & \textbf{46.7\%} & \textbf{25.7} & \textbf{82.7\%} & \textbf{26.8} \\
        \hline
        \end{tabular}
    } 
    \vspace{-2mm}
    \label{table:sota_actv_fcvid}
    \end{center} 
\end{table}
\noindent  \textbf{Results on FCVID and Mini-Sports1M.}
As shown in Table~\ref{table:sota_actv_fcvid}, our approach shows excellent efficacy and efficiency. Without additional modalities, OCSampler outperforms SCSampler by a margin of 2.4\% in mAP while using 38.8\% less computation on Mini-Sports1M and achieves 1.4\% improvement in mAP alleviating 23.6\% computational overhead over AR-Net without changing frame resolution.


\subsection{Ablation Studies}
\label{sec:exp:ablation}

\begin{table}[t]
    \centering
    \begin{footnotesize}
    \caption{\textbf{Comparisons of frame selection policies.} We report the results on different number of $N$. All of the policies use the same classifier and frame candidates, where $T$ is set to 10.}
    \vspace{-3mm}
    \label{tab:policy}
    \resizebox{\linewidth}{!}{
		\tablestyle{4pt}{1.15}
    \begin{tabular}{c|c|cccc}
    \toprule
    \multicolumn{2}{c|}{\multirow{2}{*}{Policy}} &  \multicolumn{4}{c}{mAP} \\
    \multicolumn{2}{c|}{} & \ $N=1$\  & \ $N=2$\  & \ $N=4$\  & \ $N=6$\  \\
    \midrule
    \multirow{3}{*}{\shortstack{Deterministic\\Policy}}& Random & \ 50.1\%\  & \ 62.2\%\  & \ 71.2\%\  & \ 73.8\%\  \\
    & Uniform & \ 54.2\%\  & \ 65.5\%\  & \ 72.6\%\  & \ 73.8\%\  \\
    & FrameExit & \ 54.2\%\  & \ 62.2\%\  & \ 70.4\%\  & \ 74.0\%\  \\
    \midrule
    \multirow{3}{*}{\shortstack{Learned\\Policy}}& Frame Reward & \ 61.5\%\  & \ 68.8\%\  & \ 74.2\%\  & \ 76.2\%\  \\
    & Vanilla Reward & \ 60.5\%\  & \ 69.7\%\  & \ 75.2\%\  & \ 76.6\%\  \\
    & Ours & \ \ \textbf{61.5\%}\  & \ \ \textbf{70.6\%}\  & \ \ \textbf{75.8\%}\  & \ \ \textbf{77.2\%}\  \\

    \bottomrule
    \end{tabular}}
    \end{footnotesize}
    \vspace{-3mm}
\end{table}
\noindent \textbf{Effectiveness of the learned selection policy.}
Table~\ref{tab:policy} summarizes the effect of different selection policies. For deterministic policy, we investigate three alternatives: (1) \emph{randomly} sampling frames, (2) \emph{uniformly} sampling frames, and (3) A deterministic policy proposed by \emph{FrameExit}, which can be seen as decoding videos from sparsely to densely. Besides, we also consider using different reward functions for reinforcement learning: (1) \emph{frame reward} considers the confidence of each frame rather than the integrated clip as rewards, (2) \emph{vanilla reward} removes the second item in Eq.~\ref{eq:reward} as rewards. One can observe that the learned policies have considerably better performance and the best results are obtained by our designed reward function. Notably, uniform policy appears stronger a lot than FrameExit policy when $N$ is set to 2 or 4. This is a reasonable observation, as in these cases, FrameExit policy collects more frames from the first half of videos but omits the second half while uniform policy leverages temporal information with evenly sampled frames.

\begin{table}[t]
\caption{\textbf{Effectiveness of Decision space}. The number of frame candidates $N$ is set to 6 for all settings. For $T\!=\!6$, we directly send frames to classifier without sampling.}
\vspace{-3mm}
\label{table:decision_space}
\centering
\resizebox{\linewidth}{!}{
		\tablestyle{4pt}{1.15}
\begin{tabular}{c|ccccc}
\hline
No. frame candidates & $6$ & $8$ & $10$ & $16$ & $24$\\
\hline
mAP    & 74.0\% & 76.2\% & 77.2\% & 78.0\% & 78.3\% \\ 
GFLOPs    & 24.7 & 25.6 & 25.8 & 26.4 & 27.2 \\
\hline
\end{tabular}
}
\end{table}
\noindent \textbf{Effectiveness of decision space.}
We investigate the effectiveness of decision space by using different numbers of frame candidates. As shown in Table~\ref{table:decision_space}, only adopting $T=16$ frame candidates leads to an mAP increase of 4.0$\%$ with only 1.7 GFLOPs additional computation overhead. 
An interesting phenomenon is that expanding frame candidates leads to a significant rise in accuracy performance at the beginning, but the growth gradually becomes stabilized as the candidate set becomes large, which may be attributed to the saturation of video information. In this sense, the candidate set includes salient frames to represent certain content of the video. As the expansion of candidate set, more salient frames are involved in condensing the entire video, while duplicate information might also pollute the recognition performance owing to introduced temporal redundancy. 

\begin{table}[t]
\caption{\textbf{Generality of selected frames from OCSampler}. Here we set $N$ to 4 for all classifiers. RN, MN-T and SLOW denote ResNet, MobileNetV2-TSM and SlowOnly respectively.
}
\vspace{-3mm}
\label{tab:generality}
\centering
\resizebox{\linewidth}{!}{
		\tablestyle{4pt}{1.15}
\begin{tabular}{c|ccccc}
\hline
\multirow{2}{*}{Ablation}&  \multicolumn{5}{c}{mAP(\%)} \\
  & RN & X3D-S & R(2+1)D & MN-T & SLOW\\
\hline

Baseline    & 67.5 & 62.1 & 61.1 & 57.2 & 77.1 \\ 
OCSampler   & 75.8 (\textcolor{blue}{$\uparrow$8.3}) & 68.3 (\textcolor{blue}{$\uparrow$6.2}) & 67.2 (\textcolor{blue}{$\uparrow$6.1}) & 62.0 (\textcolor{blue}{$\uparrow$4.8}) & 81.9 (\textcolor{blue}{$\uparrow$4.8}) \\
\hline
\end{tabular}
}
\vspace{-3mm}
\end{table}
\noindent \textbf{Generality of selected frames.}
These selected frames are of good generality to improve other classifiers' performance without an extra training scheduler. As shown in Table~\ref{tab:generality}, we directly apply the frames selected by OCSampler with ResNet-50 to other backbones, which also leads to significant improvements in recognition performance. 

\section{Conclusion}

In this paper, we have presented a both accurate and efficient sampling framework by condensing a video into a clip within one step, which we refer to as OCSampler. Our OCSampler avoids heavy computational overhead and addresses the problem of multiple inference times existing in most sampling methods. Moreover, our work designs a simple but reasonable reward function to consider all frames in one clip collectively rather than individually, and strikes an excellent performance on accuracy without sacrificing efficiency. We further extend our method to select adaptive numbers of frames by adopting a frame number budget module. Experiments on four widely used benchmarks verify the effectiveness of our method over existing works in terms of recognition accuracy, selection transferring, computational cost, and practical speed.


{\small
\bibliographystyle{ieee_fullname}
\bibliography{egbib}
}

\newpage
\appendix

\section*{Appendix for ``OCSampler: Compressing Videos to One Clip with Single-step Sampling''}

\section{Introduction of Prior Works}

OCSampler is compared with several competitive works that focus on efficient video recognition, including AdaFrame~\cite{wu2019adaframe}, LiteEval~\cite{wu2019liteeval}, SCSampler~\cite{korbar2019scsampler}, AR-Net~\cite{meng2020ar}, VideoIQ~\cite{sun2021dynamic}, AdaFocus~\cite{Wang_2021_ICCV}, Ada2D~\cite{li20202d}, ListenToLook~\cite{gao2020listen}, MARL~\cite{wu2019multi}, and FrameExit~\cite{ghodrati2021frameexit}.

\begin{itemize}
    \item AdaFrame~\cite{wu2019adaframe} learns to dynamically select informative frames with reinforcement learning and performs adaptive inference.
    \item LiteEval~\cite{wu2019liteeval} combines a coarse LSTM and a fine LSTM to adaptively allocate computation based on the importance of frames.
    \item SCSampler~\cite{korbar2019scsampler} introduces a light-weighted framework to efficiently identify the most salient temporal clips within a long video. We follow the implementation of \cite{meng2020ar}.
    \item AR-Net~\cite{meng2020ar} dynamically identifies the importance of video frames, and processes them with different resolutions accordingly.
    \item VideoIQ~\cite{sun2021dynamic} learns to dynamically select optimal quantization precision conditioned on input clips.
    \item AdaFocus~\cite{Wang_2021_ICCV} dynamically processes video frames with different patches accordingly.
    \item Ada2D~\cite{li20202d} learns instance-specific 3D usage policies to determine frames and convolution layers to be used in a 3D network.
    \item ListenToLook~\cite{gao2020listen} fuses image and audio information to select the key clips within a video
    \item MARL~\cite{wu2019multi} proposes to learn to select important frames with multi-agent reinforcement learning.
    \item FrameExit ~\cite{ghodrati2021frameexit} adopts a deterministic policy function and gating modules to determine the earliest exiting point for inference.
\end{itemize}

\section{Implementation Details}

In our implementation, we train $f_{\textnormal{S}}$ and $f_{\textnormal{C}}$ using an SGD optimizer with cosine learning rate annealing and a Nesterov momentum of 0.9~\cite{he2016deep,meng2020ar,lin2019tsm,Wang_2021_ICCV}. 
The size of the mini-batch is set to 64, while the weight decay is set to 1e-4. 
For ImageNet pretrained settings, we initialize $f_{\textnormal{S}}$ and $f_{\textnormal{C}}$ 
with ImageNet pretrained MobileNetV2-TSM~\cite{lin2019tsm} and ResNet-50~\cite{he2016deep}. 
For Kinetics pretrained settings, we initialize models with Kinetics-400 pretrained weight and fine-tune them on the target dataset. 
In stage I, we warm up $f_{\textnormal{S}}$ and $f_{\textnormal{C}}$ using uniformly sampled frames for 50 epochs with an initial learning rate of 0.01 and 0.005, respectively. 
In stage II, we train $\pi$ with an SGD optimizer with cosine learning rate annealing for 50 epochs and an initial learning rate of 0.001. 
We conduct all experiments on 8 TITAN XPs and will release our codes public to facilitate future works.

\section{Temporal Localization Results}

We further extend OCSampler to the temporal localization task. 
Specifically, we first use BMN~\cite{lin2019bmn} to extract action proposals and 
then use SlowOnly-R50 (which takes 8 frames as input) equipped with OCSampler to assign action labels to each proposal. 
For comparison, we also report the localization performance of using SlowOnly-8x8 
trained with fix-length sampling to assign action labels (with 10-clip testing).
Table~\ref{tab:loc_result} shows that OCSampler can achieve better localization results with far less computation consumed.


\begin{table}[h]
    \centering
    \resizebox{\linewidth}{!}{
		\tablestyle{4pt}{1.15}
	\begin{tabular}{cccccccc}
	\hline
	Methods & GFLOPs      & mAP   & AP@0.5 & AP@0.6 & AP@0.7 & AP@0.8 & AP@0.9\\
	\hline
	SlowOnly & 549   & 26.9 & 37.0  & 33.5  & 30.0  & 25.2  & 17.0  \\ \hline
	OCSampler & \textbf{68} & \textbf{28.2} & \textbf{38.8}  & \textbf{35.1}  & \textbf{31.4}  & \textbf{26.5}  & \textbf{17.8}  \\ \hline
	\end{tabular}}
	\caption{\textbf{Localization Results. } We compare the action localization performance of OCSampler and SlowOnly (fix-length sampling, 10-clip testing). OCSampler achieves superior localization performance with far less computation.}
	\label{tab:loc_result}
\end{table}

\section{The Ability of Adaptive Selection}

\begin{figure}[t]
    \centering
    \resizebox{\columnwidth}{!}{\includegraphics[width=1\textwidth]{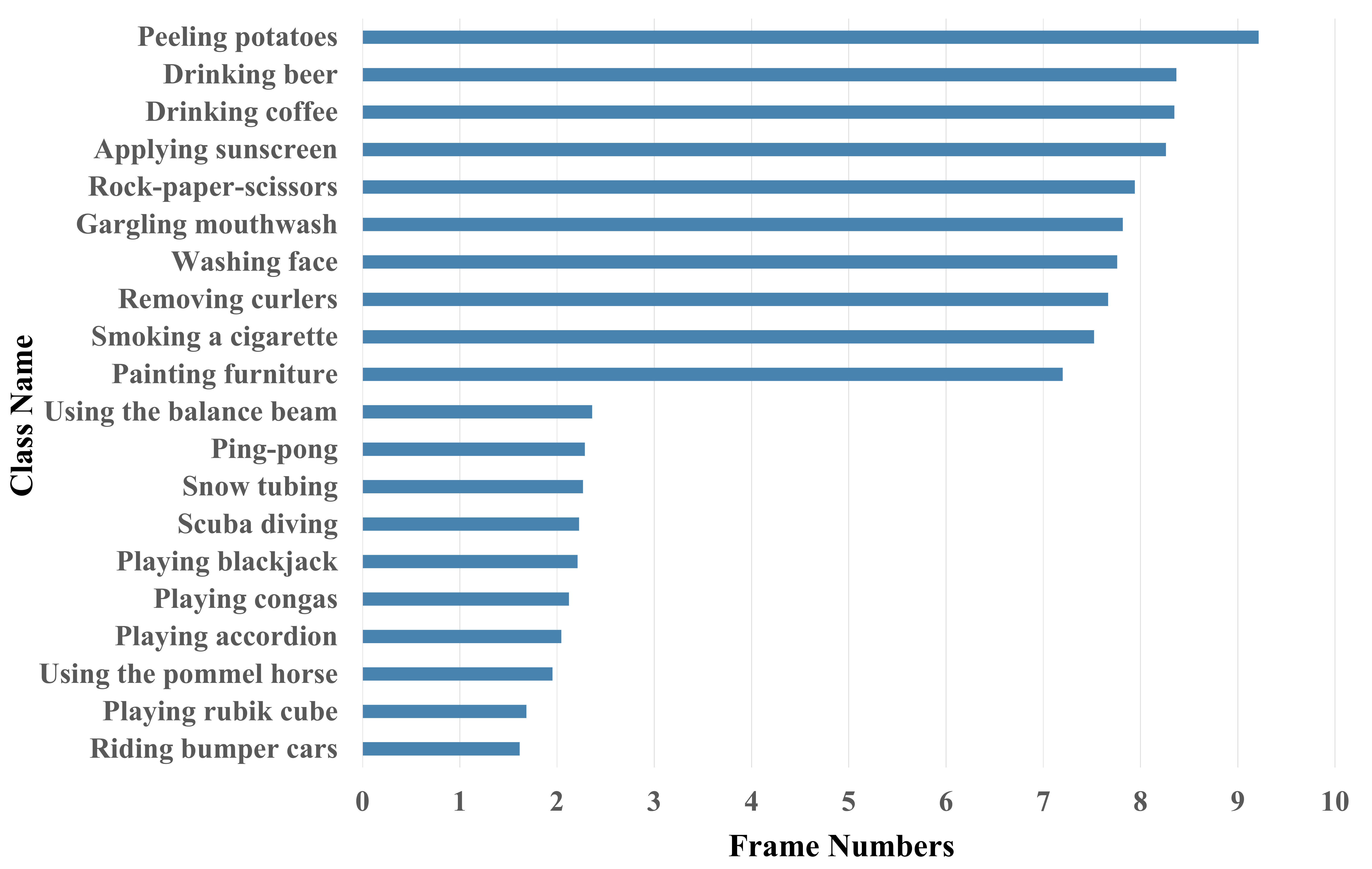}}
    \vspace{-7mm}
    \caption{\textbf{The Top-10 classes that require the most and the least number of frames in average.} Specifically, videos whose backgrounds contribute a lot demand less computational cost, while videos containing continuous and subtle actions require more frame number budgets. We visualize some cases in Figure~\ref{fig:vis}.}
    \vspace{-4mm}
    \label{fig:per_class}
\end{figure}

We statistically analyze the number of frames used in different categories. Figure~\ref{fig:per_class} shows the Top-10 classes that require the most and the least number of frames. The number of frames required by different video classes varies significantly, affected by the complexity of video content.

\begin{figure}[t]
    \centering
    \resizebox{\columnwidth}{!}{\includegraphics[width=0.8\textwidth]{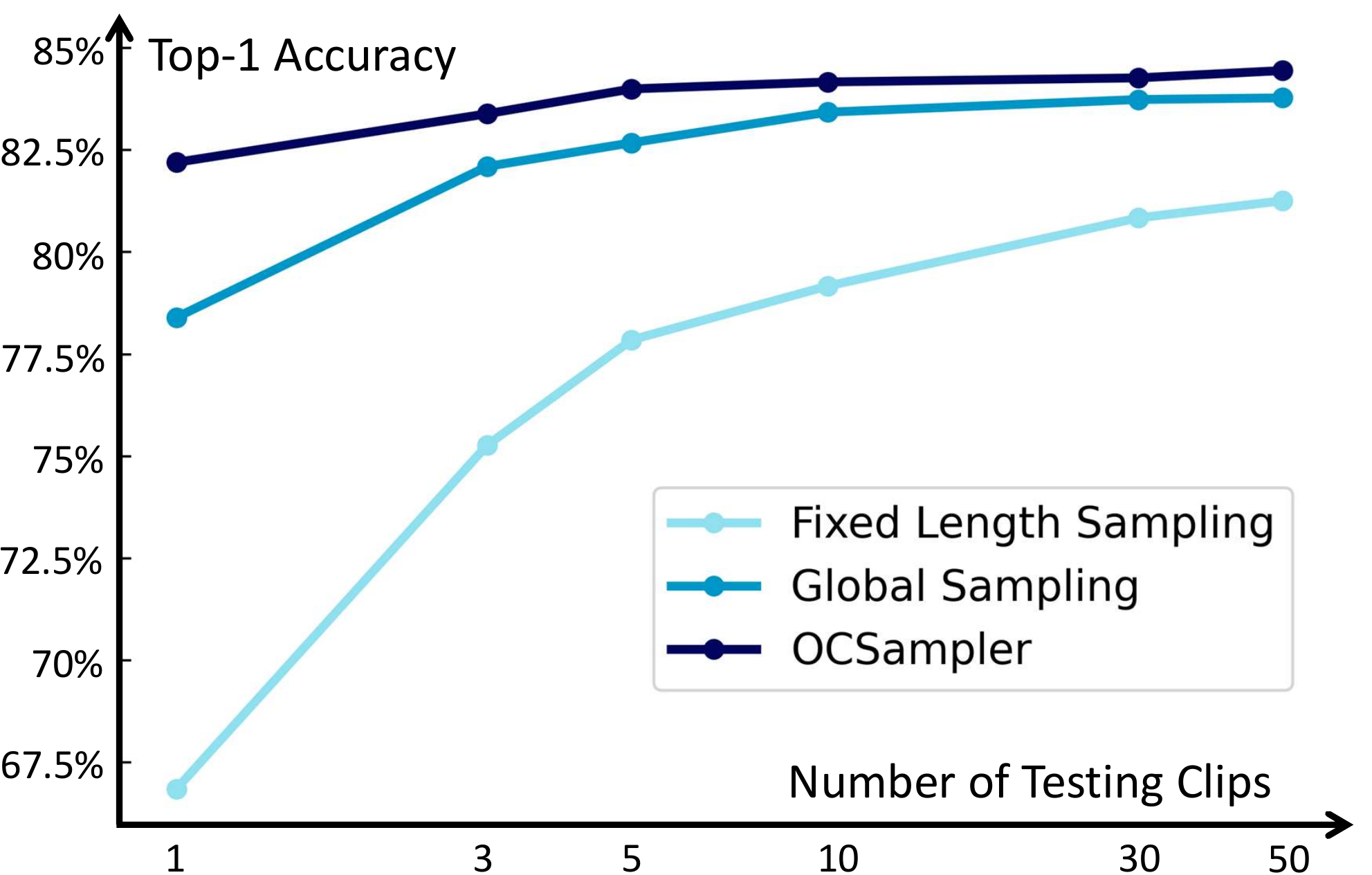}}
    \vspace{-7mm}
    \caption{\textbf{Different sampling strategies with multi-clips on ActivityNet-v1.3.} OCSampler achieves more competitive recognition performance with only one-clip testing over other strategies with multi-clip testing.}
    \vspace{-4mm}
    \label{fig:multi-clip}
\end{figure}

\begin{figure*}
\begin{center}
     \includegraphics[width=0.82\linewidth]{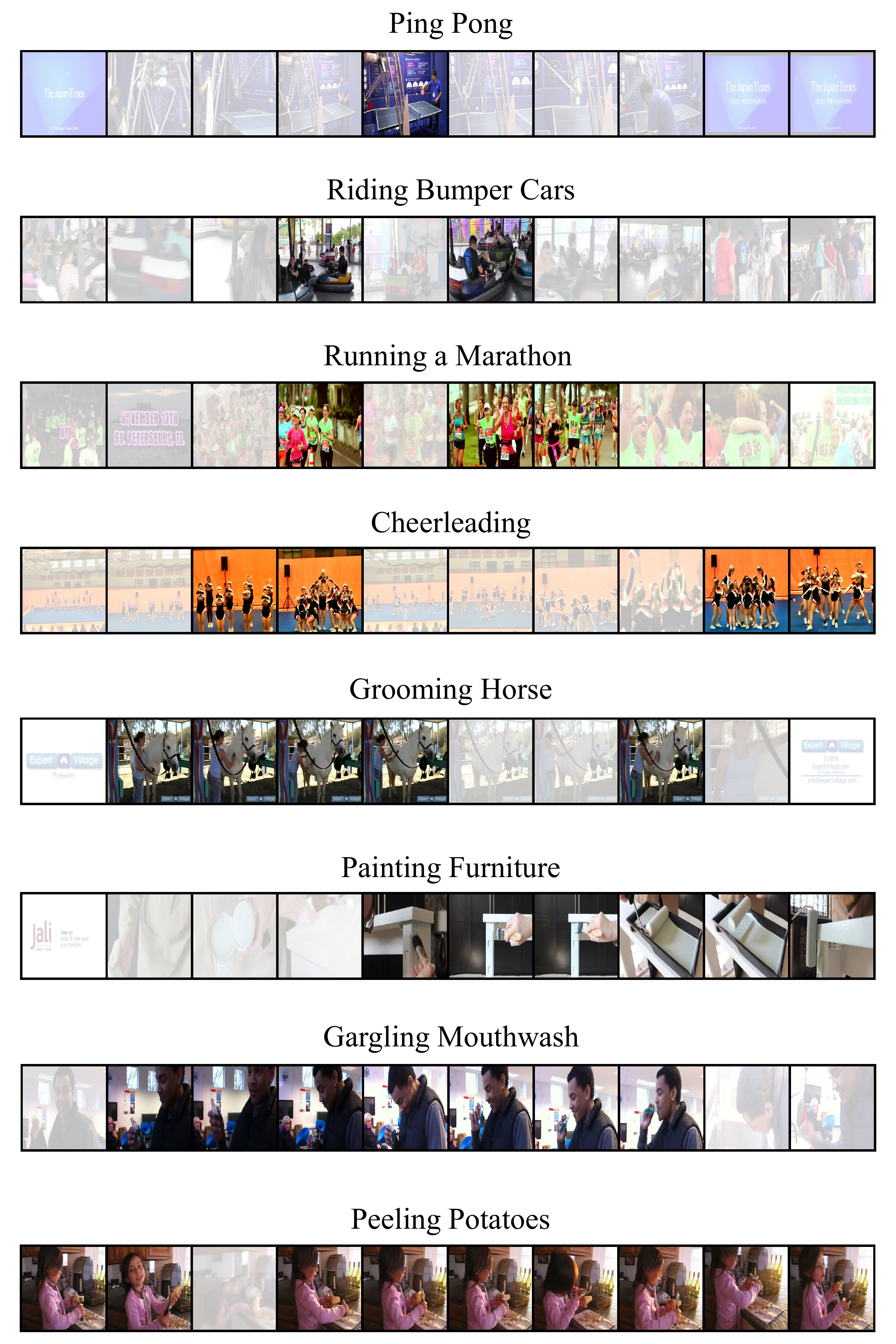}
\end{center} \vspace{-6mm}
   \caption{\textbf{Qualitative examples.} Our proposed approach \textbf{OCSampler+} processes more informative frames to form a clip for more complex videos, and takes fewer frames for simpler ones to avoid temporal redundancy and further save computational costs. Best viewed in color.}
   \vspace{-5pt}
   \label{fig:vis}
\end{figure*}

We provide additional visualization examples to illustrate the learned policy by OCSampler+ in Figure~\ref{fig:vis}. Videos are uniformly sampled in 10 frames. OCSampler+ compresses videos into one clip with informative frames, and dynamically adjusts frame number budgets for different content of videos to further reduce computational costs. Specifically, Videos whose backgrounds contribute a lot (\textit{e.g.}, "Ping Pong" and "Riding Bumper Cars" in the top 2 examples of Figure~\ref{fig:vis}) require less computational overhead, while videos containing continuous and subtle actions (\textit{e.g.}, "Gargling Mouthwash" and "Peeling Potatoes" in the bottom 2 examples of Figure~\ref{fig:vis}) take more frame number budgets for classification.

\section{Multi-Clip Results}

In this section, we compare our OCSampler using multi-clip testing with two standard sampling strategies: \emph{Fixed-Length} and \emph{Global}. \emph{Fixed-Length} samples frames only in a short temporal window to form a clip, while \emph{Global} selects frames uniformly over the entire videos. Here, we use SlowOnly-R50 with Kinetics pretrained weight on ActivityNet, and each clip is built with 8 frames. Figure~\ref{fig:multi-clip} demonstrates that OCSampler outperforms other strategies with only one clip by a large margin in recognition accuracy and efficiency.

\section{Validation with Instance-level Annotations}

\begin{figure}[t]
    \centering
    \resizebox{\columnwidth}{!}{\includegraphics[width=1\textwidth]{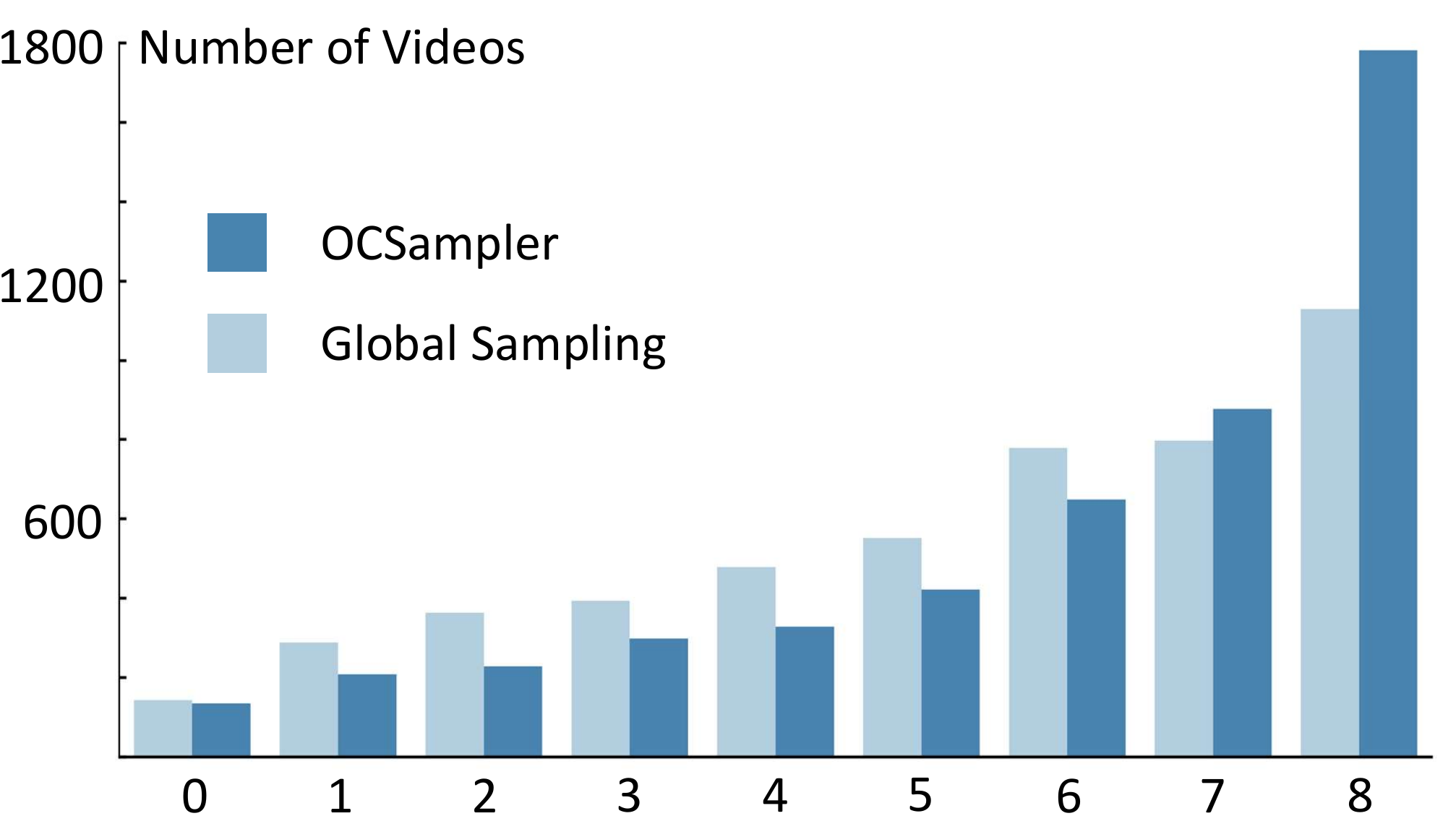}}
    \vspace{-7mm}
    \caption{\textbf{Validation with instance-level annotations.} We demonstrate how many videos have $M (0\le M\le 8)$ sampled frames in the annotated segments of ActivityNet-v1.3 validation set. OCSampler can gather more significant frames (which fall into the ground-truth segments).}
    \vspace{-4mm}
    \label{fig:instance}
\end{figure}

Besides the improved recognition performance, we find that more frames sampled by OCSampler fall into the annotated action segments compared to \emph{Global} Sampling (Figure~\ref{fig:instance}), which validates OCSampler's capability to sample informative frames from another angle. Here we set $T=32$ and $N=8$.

\section{Dataset License}

ActivityNet-v1.3~\cite{caba2015activitynet} dataset is licensed under an MIT license and Kinetics~\cite{kay2017kinetics} dataset is licensed by Google Inc. under a Creative Commons Attribution 4.0 International License. The Sports-1M~\cite{karpathy2014large} dataset is made available under a Creative Commons License.

\clearpage



\end{document}